# INDIAN SIGN LANGUAGE RECOGNITION USING MEDIAPIPE HOLISTIC

**Kaushal Goyal**
Vellore Institute of Technology
Chennai
ikaushalgoyal@gmail.com

**Dr. Velmathi G**
Vellore Institute of Technology
Chennai
velmathi.g@vit.ac.in

## ABSTRACT

Deaf individuals confront significant communication obstacles on a daily basis. Their inability to hear makes it difficult for them to communicate with those who do not understand sign language. Moreover, it presents difficulties in educational, occupational, and social contexts. By providing alternative communication channels, technology can play a crucial role in overcoming these obstacles. One such technology that can facilitate communication between deaf and hearing individuals is sign language recognition. We will create a robust system for sign language recognition in order to convert Indian Sign Language to text or speech. We will evaluate the proposed system and compare CNN and LSTM models. Since there are both static and gesture sign languages, a robust model is required to distinguish between them. In this study, we discovered that a CNN model captures letters and characters for recognition of static sign language better than an LSTM model, but it outperforms CNN by monitoring hands, faces, and pose in gesture sign language phrases and sentences.

The creation of a text-to-sign language paradigm is essential since it will enhance the sign language-dependent deaf and hard-of-hearing population's communication skills. Even though the sign-to-text translation is just one side of communication, not all deaf or hard-of-hearing people are proficient in reading or writing text. Some may have difficulty comprehending written language due to educational or literacy issues. Therefore, a text-to-sign language paradigm would allow them to comprehend text-based information and participate in a variety of social, educational, and professional settings. This research aims to improve sign language detection and employs a production system to facilitate communication between deaf and hearing individuals. Technology can enhance the lives of the deaf and facilitate their social integration.

# 1. Introduction

People who are deaf or hard of hearing have difficulties on a daily basis, particularly in communication. According to the World Health Organisation [1], by 2030, there will be 630 million people in the world, and by 2050, 900 million. India, the second-largest country on Earth, is home to some 63 million deaf or hearing impaired individuals [1]. The Indian government has been so preoccupied with modernizing the country with technological resources and infrastructure that it has ignored the requirements of deaf and hard-of-hearing citizens. A number of factors, including an aging population, environmental factors, and untreated ear infections, are expected to contribute to this increase. Due to the high cost and limited availability of hearing aids and other assistive devices, only a small fraction of individuals with hearing impairments have access to hearing aids and other assistive devices. Deaf and hard-of-hearing persons in India may benefit greatly from technological developments like sign language recognition systems in terms of accessibility and communication. Deaf and hard-of-hearing persons frequently encounter communication barriers since sign language is often the primary means of communication for many members of the deaf community[11].

Consequently, they may have difficulty engaging in social interactions, obtaining education or professional training, or gaining access to vital services and resources.
A 2011 census revealed that 73.9 percent of India's 13.4 million 15-to-59-year-olds with hearing impairment were marginal laborers [1]. This indicates that only 26.1% of the active population was employed. Thankfully, technology can play a significant role in overcoming these obstacles. Speech-to-text innovations, such as real-time transcription and automatic closed captioning, can provide deaf and hard-of-hearing individuals with greater access to spoken language, allowing them to partake in dialogues and meetings more effectively. In addition, sign language recognition systems that use machine learning algorithms to translate sign language into spoken or written language can assist in bridging the communication gap between the deaf and hearing communities.
In this paper, we explore the various technologies available to assist deaf and hard-of-hearing individuals, highlighting their potential to enhance communication and access to services and resources for Indian Sign Language. We also investigate the challenges associated with the development and deployment of these technologies, including precision, dependability, and accessibility concerns. In conclusion, we argue that continued investment in research and development is required to ensure that technology can meet the requirements of the deaf and hard-of-hearing community and promote greater inclusion and accessibility for all. As shown in Figure 1, we have trained our model-based Indian Sign Language (ISL) character.

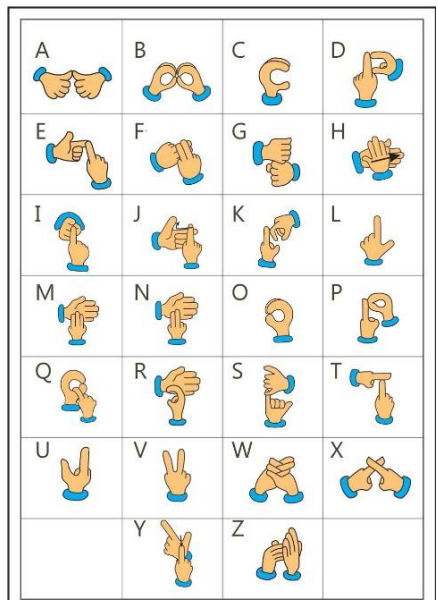

Fig 1. Indian Sign Language Alphabets [24]

In addition to recognizing sign language, we also convert text to Indian Sign Language, as many users of sign language may have difficulty understanding the text. For those who are deaf, hard of hearing, or have no knowledge of sign language, this approach makes communication easier. By doing this, a solid structure is created for the deaf and hard-of-hearing populations, which is crucial for effective communication on both sides.

## 2. Background

Sign language is a complex and facially expressive mode of communication. The fact that many hearing persons are not proficient in sign language, despite the fact that it is an essential mode of communication for the deaf and hard of hearing[11], can greatly impede communication. People who are hard of hearing or deaf may experience challenges that limit their ability to fully participate in society. To address this issue, researchers have been working on automatic ways to translate sign language into text or voice. These technologies may alter how people who are hard of hearing and deaf interact with the hearing world. The quality of life for the deaf and hard of hearing can be greatly enhanced by these devices' real-time sign language translation.

The Indian community of deaf and hard-of-hearing uses Indian Sign Language (ISL), to communicate with each other and with hearing individuals. ISL, unlike spoken languages, conveys meaning through a combination of hand gestures, facial expressions, and body language. ISL's complex history dates back to the early 19th century, when the British used it to communicate with deaf individuals in India[20]. ISL has developed over time into a distinct and intricate language with its own grammar, syntax, and vocabulary. One of the most widely used sign languages in the world, ISL is thought to be used by over a million deaf people in India. It is impossible to overstate the importance of Indian Sign Language to the deaf people in India as a way of communication. It facilitates access to education, healthcare, employment, and other essential services for deaf individuals. Nevertheless, despite its ubiquitous use, ISL recognition remains difficult due to the complexity and variability of sign language gestures[11]. Consequently, more precise and efficient methods for recognizing ISL gestures are required, which could have a significant impact on the lives of deaf individuals in India.

The current state of the art in ISL recognition employs computer vision and machine learning techniques to recognize sign language gestures from video data automatically. Using computer vision algorithms to monitor the movements of the hands and other body parts in the video and extract features that can be used to recognize specific signs is one of the most common approaches. Then, machine learning algorithms, such as neural networks, can be trained on these features to learn how to identify various signals in the video data. In some instances, these methods have yielded promising results, but they also have some limitations. One difficulty is the high variability and complexity of sign language gestures, which can make it difficult to recognize signs accurately in various contexts and by different signers. The absence of standardized datasets and evaluation metrics makes it difficult to compare and evaluate the performance of different ISL recognition methods. In addition, many existing methods require significant amounts of annotated data to train machine learning models, which can be costly and time-consuming to collect. Recent advances in computer vision and machine learning are advancing ISL recognition and paving the way for more accurate and efficient methods in the future, despite these obstacles.

## 3. Related Work

In the paper Development of Sign Language Motion Recognition System for Hearing-Impaired People Using Electromyography Signal, a real-time motion recognition system based on an electromyography signal was proposed for recognising actual ASL hand motions in order to help hearing people learn sign language and to assist deaf people in communicating with others (Tateno et al., 2020)[2]. A hand glove system that can convert sign language motions into letters shown on an LCD screen is one example of recent research employing sensors. Another example is the sign to text translation system using a hand glove. Deaf and mute persons can use this approach to communicate with those who don't know sign language. Flex resistors, a microprocessor, and an LCD display make up the system.It aims to overcome communication barriers and ensure effective message transmission. (Hayek et al, 2014) [4].

While there have been very notable research made on this topic, using various machine learning techniques for hand detection and tracking are relatively newer [3][11]. Real-time Indian Sign Language (ISL) Recognition tracks features usign a grid-based framework technique that transforms it into a feature vector, hand poses are then classified using the k-nearest neighbors algorithm (Shenoy et al, 2018)[11] [3]. Another research, Sign Language Recognition Using Two-Stream Convolutional Neural Networks with Wi-Fi Signals (Lee et al, 2020) provides a Wi-Fi-based sign language recognition approach that preprocesses Channel State Information using singular value decomposition. The method combines spatial and motion streams using a convolutional neural network and includes a mechanism to choose key features. On the SignFi dataset, the technique showed excellent recognition accuracy rates, suggesting superior performance and generalization capabilities when compared to competing methods [5].

In Continuous sign language recognition: Towards large vocabulary statistical recognition systems handling multiple signers (Koller et al, 2006) [6] they put forth a method that may be used in the actual world to identify signers using a large vocabulary of sign language.They focus on five areas and conduct experiments to show the value of tracking, multimodal sign language elements, and temporal derivatives in the identification of sign language. They look at visual modelling techniques and deal with signer dependence using class language models and CMLLR adaptation. On two publicly accessible massive vocabulary databases reflecting lab data and unrestricted real-life sign language, the offered technique yields low word error rates. The big vocabulary databases that are freely accessible are the "SIGNUM database, which has 25 signers, 455 sign words, 19k phrases, and unrestricted real-life sign language" ([6,7]). Word error rates of up to 10.0%/16.4% and, respectively, up to 34.3%/53.0% for single signer/multi-signer configurations are achieved using the "RWTH-PHOENIX-Weather database (9 signers, 1081 sign vocabulary, 7k sentences)".

Camgoz et al. (2018) introduced the Sign Language Translation (SLT) challenge for the first time. It aims to translate sign language films into spoken English while taking into account sign language's complex grammatical and syntactic patterns. As they formalised SLT inside the Neural Machine Translation (NMT) framework, the mapping between sign language and spoken language, the spatial representations, and the language model were all jointly trained. In order to evaluate the effectiveness of their Neural SLT, they gathered the first "Continuous SLT dataset, RWTH-PHOENIX-Weather 2014T", which consists of more than 67K signs from a sign vocabulary of more than 1K and more than 99K words from a German vocabulary of more than 2.8K. Their end-to-end tokenization networks at the frame- and gloss-levels scored 9 and 18, respectively. [7][11].

The Indian Sign Language recognition method makes use of SURF, SVM, and CNN. Katoch et al. (2022) proposed a technique for detecting Indian Sign Language alphabets and numbers using the [9] Bag of Visual Words model (BOVW), SURF (Speeded Up Robust Features) features, and classification models like Support Vector Machines (SVM) and Convolutional Neural Networks (CNN). Hand motions may be separated using the background removal and skin tone techniques. To link the obtained attributes to the relevant labels, histograms are employed. The system generates predicted labels in text and audio using Pyttsx3 and the Google voice API. The proposed method appears to be successful in recognising ISL alphabets and numerals, making it a valuable tool for speech-impaired persons who require communication.

Finally, the study of Indian Sign Language (ISL) recognition has gotten a lot of attention from scholars in recent years because of its potential to improve communication and access to information for India's deaf and hard-of-hearing community. Various techniques such as computer vision, machine learning, and deep learning have been used to construct accurate and efficient ISL recognition systems, according to the related research reviewed in this paper. However, further research is needed to solve the problems that make ISL detection difficult, such as the diversity of gestures, lighting circumstances, and intricate hand movements. Overall, the results of the associated studies point to the possibility of ISL recognition systems to enhance the deaf and hard-of-hearing community's quality of life, and more study in this area may result in important advances.

## 4. Research gap
The research described in the related work addresses the need for a more precise and effective Indian Sign Language (ISL) recognition system. Despite the increasing number of studies on sign language recognition,

ISL recognition research remains scarce. A review of the existing literature reveals that the majority of studies have focused on American Sign Language (ASL) or other sign languages, such as British Sign Language (BSL), Australian Sign Language (Auslan), and Chinese Sign Language (CSL). Few studies have examined the recognition of ISL, and the performance of existing systems is frequently unsatisfactory due to the complexity of ISL and the lack of a comprehensive dataset.

Traditional methods rely on high computational data by processing videos which require huge storage of datasets and even more advanced systems for model training. A simpler approach and easy-to-train models are thus required to enhance the real-time detection of Indian Sign Language (ISL). The methodology proposed in this system can be used by anyone to train on their own facial expressions and hand gestures thus reducing the chances of inaccurate classification and providing a robust system for the detection and translation of sign language.

As proposed, our system extracts only the key points and saves them in a numpy array, thus reducing the size of the dataset and faster processing of models. We eliminate the use of image operations and storage of data-intensive videos, making the model much simpler and easy to recognize sign language.

The research void in this field is therefore the development of a more robust and accurate ISL recognition system that can effectively recognize both static and gesture sign languages and that can be tested and evaluated using a comprehensive dataset of ISL. This would enhance the communication skills of deaf and hard-of-hearing individuals who rely on ISL and promote social inclusion within this community.

## 5. Methodology

We manually extracted frames from sign language videos to generate the static and gesture sign language datasets. For the static sign language dataset, we focused on capturing hand positions and gestures, whereas for the gesture sign language dataset, we captured the entire sign language phrase, including full body movements, facial expressions, and hand gestures. We analyzed the static dataset of sign language using a CNN model with three convolutional layers. Flattening the final output before feeding it to a fully connected layer with ReLU activation and a sigmoid classification output layer. We utilized the sparse categorical cross-entropy loss function and Adam optimizer for training. We employed the same training parameters for the dataset of gesture sign language.

We employed an LSTM model with three LSTM layers, each followed by a batch normalization layer, to avoid overfitting the static sign language dataset. The output is input into two fully connected layers with ReLU activation, a dropout layer, and a sigmoid output layer for classification. We utilized the same loss function and optimizer as CNN's model. We used an LSTM model with three LSTM layers and no batch normalization for the gesture sign language. We utilized the same classification layers along with a softmax output layer in place of a sigmoid layer.

We utilized the same loss function and optimizer as in the case of the static sign language model. For reproducibility, we trained both models on the same dataset, which was split into training and testing sets with a 20% test size and a 42-sample random sample. Both models were trained with optimal epochs, and validation accuracy was monitored to avoid overfitting. Several evaluation metrics were utilized to assess the performance of the proposed system. In addition, the efficacy of CNN and LSTM models on static and gesture sign language datasets was compared.

Overall, the proposed system achieved a high degree of precision and performed well in sign language recognition tasks. The CNN model performed better than the LSTM model on the static sign language dataset, whereas the LSTM model performed flawlessly on the gesture sign language dataset, demonstrating the versatility of both models.

### 5.1 Dataset Collection

The first stage of this analysis is to collect the necessary data for training and evaluating the models. Using MediaPipe Holistic and OpenCV in conjunction with a webcam, we captured video data of individuals performing sign language gestures. MediaPipe Holistic provides pose, hand, and face landmark estimation, which assists in identifying sign language hand and body movements.

The data will be collected in real-time using a predefined set of sign language gestures. The video data was not saved in order to conserve storage space. Instead, we simply stored an array of the extracted key elements from MediaPipe Holistic.

**Video Input:** MediaPipe Holistic accepts video from a camera or a previously recorded file.

**Pose Estimation:** The first step in using MediaPipe Holistic is to detect the person's pose within the video frame. This entails locating important body components such as the head, shoulders, arms, hips, and legs. MediaPipe Holistic performs this task using the BlazePose machine learning model.

**Hand and Face Landmark Estimation:** Once the pose has been estimated, the next stage is to identify the facial and hand landmarks. MediaPipe Holistic employs distinct machine-learning models for the estimation of hand and visage landmarks. The face landmark model detects 468 key points on the visage, whereas the hand landmark model detects 21 key points on each hand.

**Integration of Keypoints:** After identifying the landmarks on the hands, face, and body, MediaPipe Holistic combines the key points to create a holistic representation of the individual in the video frame. Combining the key points into a single coordinate space enables accurate monitoring of the person's movements.

**Output:** MediaPipe Holistic generates a set of key points or landmarks representing the pose, hand, and face of the individual in the video frame.

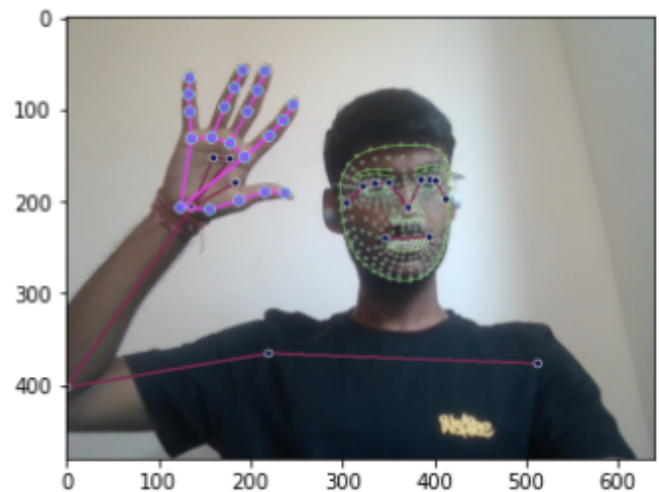

Fig 5.1.1 - Sample image of the dataset collection process

MediaPipe Holistic is used to monitor key points by analyzing video frames and identifying a person's pose, hands, and face landmarks. It combines these key points to create a comprehensive representation of the individual in the video frame, enabling precise tracking of their movements.

For each frame, a total of 1662 key points based on pose, hand, and face landmarks were collated in the form of a single coordinate space. If any of these points were not monitored by the model, we passed zero values to preserve the array's shape.

```
In [285]: np.array(sequences).shape
Out[285]: (1560, 30, 1662)
```

Fig 5.1.2 - Size of array

For this study, we divided our data acquisition into two categories: static and moving signs. For each sign, 60 videos were captured, with each video consisting of 30 frames, and for each frame, 1662 key points were captured to be stored in an array.

As we intended to divide our dataset into two categories, static and gesture, we sorted our data. First, we captured videos of alphabets, which are predominantly inert, and then we captured videos of words and phrases that required gestures.

Then, we developed a label map for the sequences that will be utilized to train the dataset.

```
In [302]: print("Label Map:", ", ".join([f"{label}:{num}" for label, num in label_map.items()]))
          Label Map: a:0, b:1, c:2, d:3, e:4, f:5, g:6, h:7, i:8, j:9, k:10, l:11, m:12, n:13, o:14, p:15, q:16, r:17, s:18, t:19, u:20,
          v:21, w:22, x:23, y:24, z:25

In [49]: label_map

Out[49]: {'hello': 0, 'thanks': 1, 'iloveyou': 2, 'indian': 3, 'how_are_you_doing': 4}
```

Fig 5.1.3 - Label map

## 5.2 Model

Model training is a crucial step in the development of a system for sign language recognition using machine learning techniques. This section will discuss the training methodology utilized for the models utilized in this project. To train the model on two distinct datasets and compare the use of time series models like LSTM versus a CNN mode model, which is better suited for image processing.

Before training our machine learning models, we preprocessed the data to ensure its suitability for model training. The preprocessing steps included extracting the key points from each frame of the videos and resizing them to a standard array size. We also divided the datasets into 80:20 training and testing sets to ensure that our models could accurately predict the sign language gestures

**CNN Model:**
The model is trained using a dataset of sequences of sign language. Using the function train test split from the SKLEARN library, the dataset is divided into training and testing sets.

Looping through each action and sequence in the dataset and appending the feature arrays to a window array generates the sequences. The window array is then appended to the sequences array, followed by the label. This is performed to generate training data. The feature arrays are generated using media pipe holistic and OpenCV to monitor the hand's key points, which are then saved as.py files. These.npy files are then imported and concatenated to form each frame's feature array. The resultant feature array is added to the window array.

The data is reshaped into a 4D array for use as input to the Convolutional Neural Network (CNN) model following the creation of the training and assessment data. Multiple layers comprise the CNN model, including Conv2D layers, MaxPooling2D layers, flattening layers, dense layers, and dropout layers. Compiling the model with the compile function specifies the optimizer, loss function, and metrics to be used during training evaluation. Because the labels are integers and not one-hot encoded vectors, the sparse categorical cross-entropy loss function is utilized.

The model is trained to utilize the fit function, with the number of epochs and validation data specified. Backpropagation with stochastic gradient descent is used to train the model, and during each epoch, the weights are adjusted to minimize the loss function. The accuracy of the model is then evaluated using the test set.

**LSTM model:**
The first layer of the model is a 64-unit LSTM layer that accepts 30 timesteps and 1662 features as input sequences. ReLU is the activation function for the LSTM layer. The return sequences parameter is set to True, indicating that the component will return sequences as opposed to the last output. The input shape parameter specifies the input data's shape. Bulk normalization is performed following each LSTM layer. This reduces the effect of vanishing gradients and accelerates the network's convergence during training. After each batch normalization layer, a dropout layer with a rate of 0.2 is implemented to prevent overfitting.

The kernel regularizer parameter applies L2 regularisation to the kernel weights of the second LSTM layer, which consists of 128 units. The third LSTM layer contains 64 units, but the return sequence is set to False, so it only returns the sequence's ultimate output. Adding two dense layers with ReLU activation functions to the model. The first dense layer has 64 units, while the second dense layer has 128 units. Each dense layer has a kernel regularizer parameter for L2 regularisation.

The final dense layer has 26 units, which corresponds to the number of classes in the dataset. The activation function of this layer is sigmoid because the assignment is a classification problem involving multiple classes. Using the Adam optimizer and a learning rate of 0.001, the model is generated. Since this is a classification problem involving multiple classes, the loss function employed is categorical cross-entropy. During training, accuracy is the metric used to evaluate the model. Callbacks are used to undertake specific actions at specific points during training. A learning rate scheduler is used to modify the optimizer's learning rate. Training is terminated using an early halting response if the model's accuracy on the validation set does not improve after a predetermined number of epochs. TensorBoard notifications are also utilized to document the training process and display the results in TensorBoard.

Using the fit method of the model object and the training data and labels as inputs, the model is trained. Along with the previously defined callbacks, the validation data and labels are also provided as inputs. Seventy epochs are utilized for instruction.

### 5.3 Model Evaluation

We determined the efficacy of the models on the evaluation set by measuring their precision. We also created confusion matrices to analyze the performance of the model in greater depth. We contrasted the performance of the models to determine which model for each dataset performed the best.

```
Model: "sequential_37"
_________________________________________________________________
Layer (type)                 Output Shape              Param #
=================================================================
conv2d_21 (Conv2D)           (None, 28, 1660, 64)      640

max_pooling2d_11 (MaxPoolin  (None, 14, 830, 64)       0
g2D)

conv2d_22 (Conv2D)           (None, 12, 828, 128)      73856

max_pooling2d_12 (MaxPoolin  (None, 6, 414, 128)       0
g2D)

conv2d_23 (Conv2D)           (None, 4, 412, 64)        73792

max_pooling2d_13 (MaxPoolin  (None, 2, 206, 64)        0
g2D)

flatten_6 (Flatten)          (None, 26368)             0

dense_95 (Dense)             (None, 64)                1687616

dropout_89 (Dropout)         (None, 64)                0

dense_96 (Dense)             (None, 26)                1690

=================================================================
Total params: 1,837,594
Trainable params: 1,837,594
Non-trainable params: 0
_________________________________________________________________
```

```
Model: "sequential_59"
_________________________________________________________________
Layer (type)                 Output Shape              Param #
=================================================================
lstm_111 (LSTM)              (None, 30, 128)           916992

batch_normalization_128 (Ba  (None, 30, 128)           512
tchNormalization)

lstm_112 (LSTM)              (None, 30, 256)           394240

batch_normalization_129 (Ba  (None, 30, 256)           1024
tchNormalization)

lstm_113 (LSTM)              (None, 256)               525312

batch_normalization_130 (Ba  (None, 256)               1024
tchNormalization)

dense_147 (Dense)            (None, 256)               65792

batch_normalization_131 (Ba  (None, 256)               1024
tchNormalization)

dense_148 (Dense)            (None, 128)               32896

batch_normalization_132 (Ba  (None, 128)               512
tchNormalization)

dropout_123 (Dropout)        (None, 128)               0

dense_149 (Dense)            (None, 26)                3354

=================================================================
Total params: 1,942,682
Trainable params: 1,940,634
Non-trainable params: 2,048
_________________________________________________________________
```

(a) - Model summary for CNN for static dataset  (b) - Model summary for LSTM for static dataset

Fig 5.3.1 - Static Sign Language Dataset

```
Model: "sequential_11"
_________________________________________________________________
Layer (type)                 Output Shape              Param #
=================================================================
lstm_33 (LSTM)               (None, 30, 64)            442112

lstm_34 (LSTM)               (None, 30, 128)           98816

lstm_35 (LSTM)               (None, 128)               131584

dense_32 (Dense)             (None, 128)               16512

dense_33 (Dense)             (None, 64)                8256

dense_34 (Dense)             (None, 5)                 325

=================================================================
Total params: 697,605
Trainable params: 697,605
Non-trainable params: 0
_________________________________________________________________
```

```
Model: "sequential_19"
_________________________________________________________________
Layer (type)                 Output Shape              Param #
=================================================================
conv2d_21 (Conv2D)           (None, 28, 1660, 32)      320

max_pooling2d_21 (MaxPoolin  (None, 14, 830, 32)       0
g2D)

conv2d_22 (Conv2D)           (None, 12, 828, 64)       18496

max_pooling2d_22 (MaxPoolin  (None, 6, 414, 64)        0
g2D)

conv2d_23 (Conv2D)           (None, 4, 412, 32)        18464

max_pooling2d_23 (MaxPoolin  (None, 2, 206, 32)        0
g2D)

flatten_7 (Flatten)          (None, 13184)             0

dense_49 (Dense)             (None, 32)                421920

dense_50 (Dense)             (None, 5)                 165

=================================================================
Total params: 459,365
Trainable params: 459,365
Non-trainable params: 0
_________________________________________________________________
```

(a) - Model summary for LSTM for gesture dataset  (b) - Model summary for CNN for gesture dataset

Fig 5.3.2 - Gesture Sign Language Dataset

We studied different variations for both models and evaluated the accuracy and F1 score for both models, even though the LSTM model can be found to have good accuracy scores on train data of static signs when we put it through test data it shows an F1 score of 0.8557 compared to what we get from CNN model of 0.9775.

```
10/10 [==============================] - 2s 216ms/step
10/10 [==============================] - 2s 204ms/step
Model LSTM Accuracy: 0.8557692307692307
Model LSTM F1 score: 0.8474237708821871
Model LSTM Precision:  0.8695013783036972
Model LSTM Recall:   0.8557692307692307
```

Fig 5.3.3(a) - Evaluation of LSTM model for the static dataset

```
10/10 [==============================] - 6s 552ms/step - loss: 0.1461 - accuracy: 0.9776
Test loss: 0.146091580390930018
Test accuracy: 0.9775640964508057
10/10 [==============================] - 6s 547ms/step
Model CNN Accuracy: 0.9775641025641025
Model CNN F1 score: 0.9772205016112788
Model CNN Precision:  0.9809981684981685
Model CNN Recall:   0.9775641025641025
```

Fig 5.3.3(b) - Evaluation of CNN model for the static dataset

Even though LSTM outperforms CNN on a static sign dataset, the LSTM model proves to be more effective when we evaluate gesture-based signs. It shows complete 100% accuracy for the LSTM model compared to 90% from the CNN model.

```
Accuracy: 1.0
F1 score: 1.0
Model LSTM Precision: 1.0
Model LSTM Recall: 1.0
```

Fig 5.3.4(a) - Evaluation of LSTM model for gesture dataset

```
Model CNN Accuracy: 0.9
Model CNN F1 score: 0.8975589225589224
Model CNN Precision: 0.9257142857142857
Model CNN Recall: 0.9
```

Fig 5.3.4(b) - Evaluation of CNN model for gesture dataset

Our research highlights the importance of selecting appropriate models for different categories of sign language recognition tasks and provides insight into the effectiveness of various model architectures for this purpose. Overall, we aimed to provide a comprehensive evaluation of the proposed sign language recognition system and to compare the performance of different models for different types of sign language recognition tasks.

## 6. Implementation of the model and algorithms

To train our model, our research on sign language recognition required the development of an exhaustive dataset of sign language gestures. To accomplish this, we concentrated on static signs with a fixed hand shape and compiled a dataset of 26 static signs, each of which was designated with the alphabet class to which it corresponds, and for gesture-based signs, we took 5 words to train our model.

To capture the sign language gestures, we used OpenCV with a webcam and media pipe holistic to monitor and display the key points of the hand gestures for enhanced recognition. This allowed us to store the resulting key points for each class and video in an array, which was then used to train our sign language recognition model.

After training the model, we incorporated it into a Django web application with two sections for sign language communication: one for converting sign language to text and another for converting text to sign language. We used a speech recognition API that can separate spoken words into individual components to convert speech to text. If the spoken term exists in our database, we convert it to sign language and display it on the website. If the word is not in the database, the API separates it into individual letters and converts each letter to its corresponding sign language display.

By combining openCV, media pipe holistic, CNN, and a speech recognition API, we were able to develop a highly accurate and reliable system for sign language recognition and conversion, which has the potential to improve communication between the deaf and hard-of-hearing community and the rest of the world.

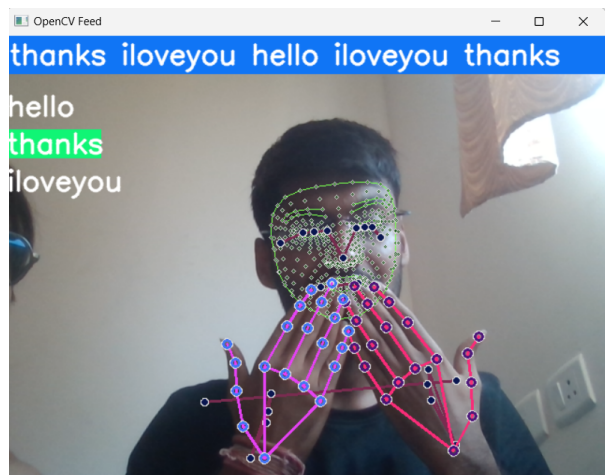

Fig 6.1 - Live feed from a webcam while detection

# 7. Results

In our research on the recognition of Indian sign language, we classified static and gesture sign language gestures using two distinct datasets. Individual alphabet signs were classified using LSTM and CNN models for static sign recognition, while complete sentences were identified using an LSTM model for gesture sign recognition. According to our research, using sequence-based models for sign recognition based on time series prediction differs significantly from using CNN models for processing key points in 3D space.

In particular, our findings demonstrated that LSTM models were more effective at recognizing gesture sign language gestures, whereas CNN models were more effective at recognizing static sign language gestures.

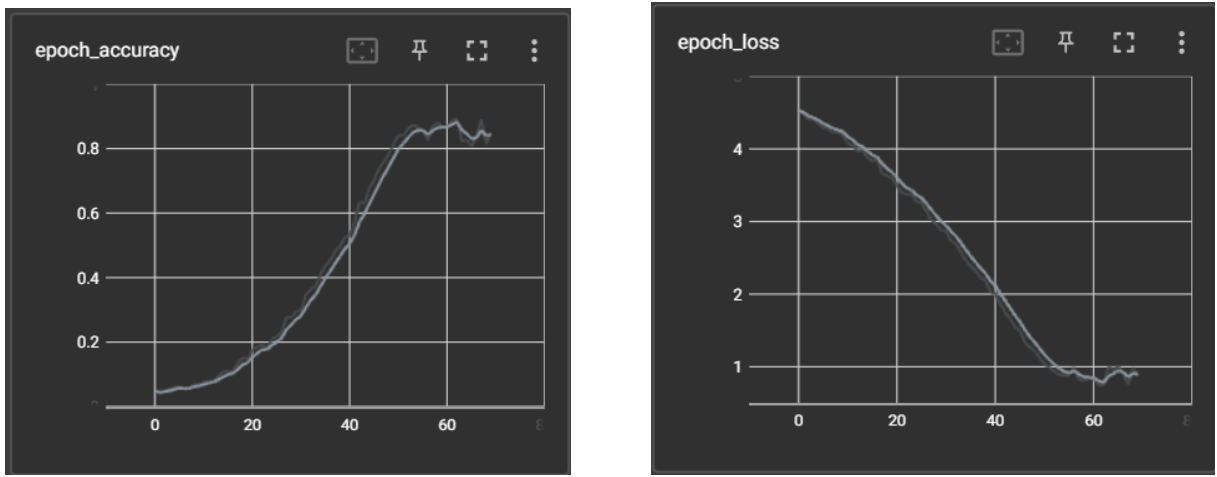

Fig 7.1 - Tensorboard graphs for accuracy and loss through the epochs

In the context of Indian Sign Language recognition, attaining a train data accuracy of 85 percent is a noteworthy accomplishment. This level of precision is attributable to the efficient use of LSTM and CNN models for recognizing static and dynamic sign language gestures, respectively. An important indicator of the model's ability to learn and generalize patterns from the data is the decrease and tapering off of the loss during training. It indicates that the model is no longer learning from the training data and is capable of accurately predicting the test data.

In addition, the optimal number of iterations required to attain the highest level of accuracy can vary depending on the dataset's complexity and size. In the case of ISL recognition, our research indicates that training the dataset for sixty to seventy iterations can yield the greatest results. This emphasizes the significance of fine-tuning model parameters and optimizing the training process to obtain the best results possible.

| Model | Accuracy | F1-score | Precision | Recall |
|---|---|---|---|---|
| LSTM | 0.856 | 0.847 | 0.870 | 0.856 |
| CNN | 0.978 | 0.978 | 0.981 | 0.978 |

Fig 7.2(a) - Comparison results of both models for static signs

| Model | Accuracy | F1-score | Precision | Recall |
|---|---|---|---|---|
| LSTM | 1.0 | 1.000 | 1.000 | 1.0 |
| CNN | 0.9 | 0.898 | 0.926 | 0.9 |

Fig 7.2(b) - Comparison results of both models for gesture signs

Our findings revealed that the use of sequence-based models for sign recognition based on time series prediction differs significantly from the use of CNN models for processing key points in 3D space. Specifically, we found that LSTM models were more effective at recognizing gesture sign language, whereas CNN models were more effective at recognizing static sign language.

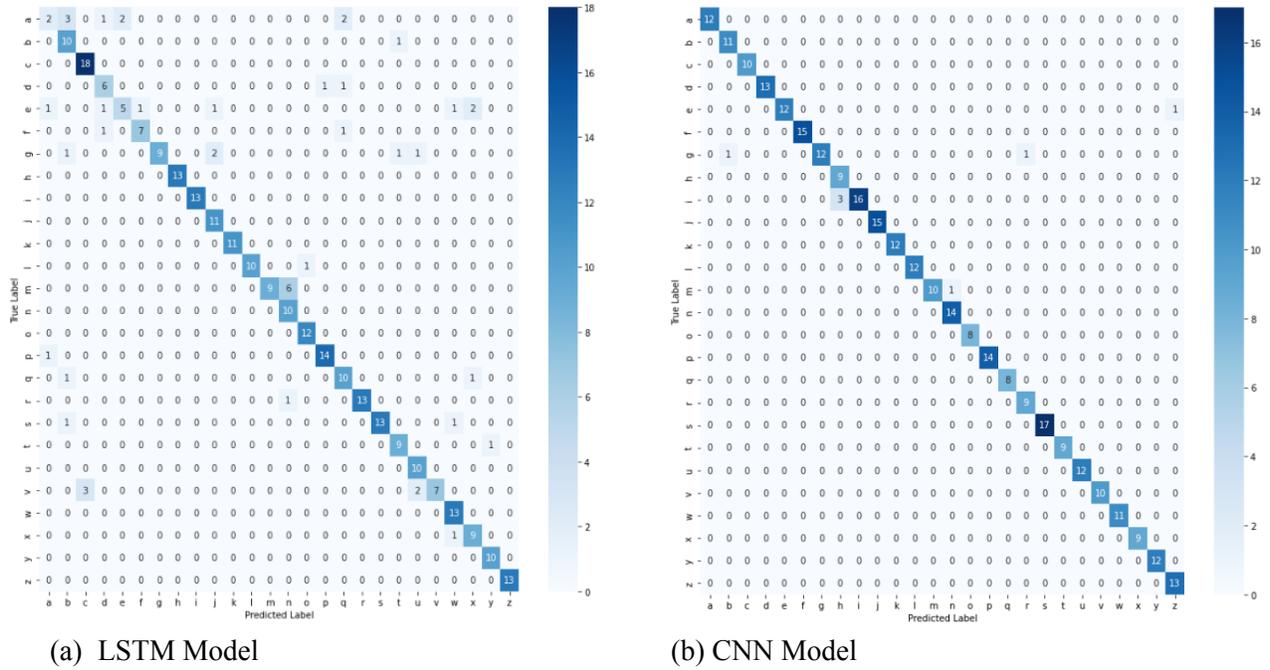

(a) LSTM Model                    (b) CNN Model

Fig 7.3 - Confusion matrix for static sign language dataset

For gesture sign language recognition using the LSTM model, we also calculated the accuracy, precision, recall, and F1 score. Additionally, we used the confusion matrix to visualize the performance of the model. The confusion matrix allows for a deeper analysis of its performance.

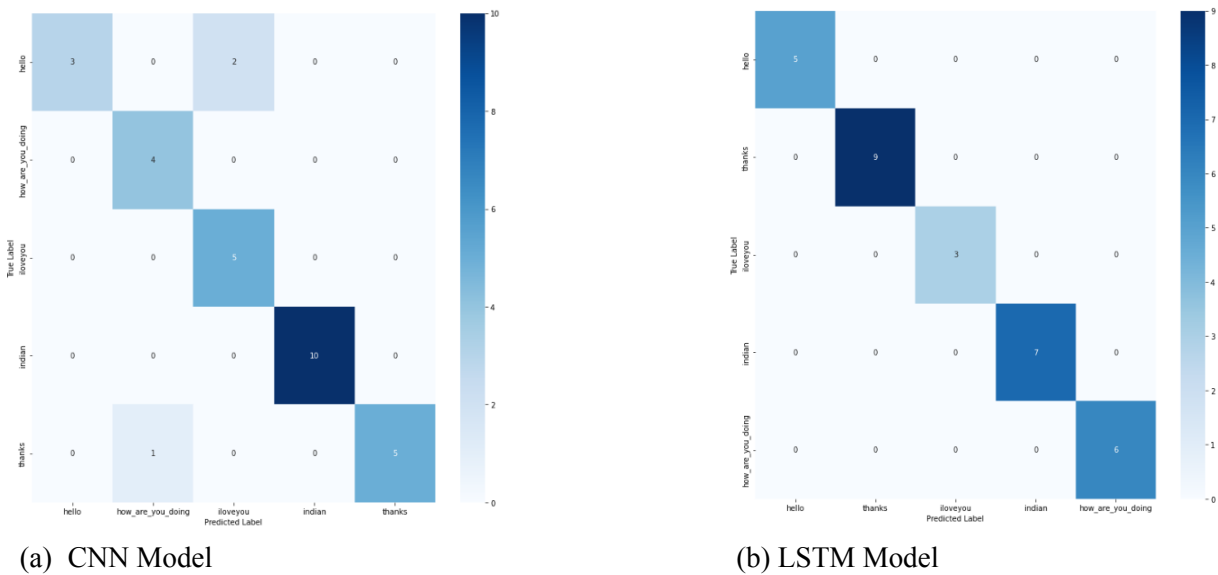

(a) CNN Model                    (b) LSTM Model

Fig 7.3 - Confusion matrix for gesture sign language dataset

Thus we saw the different evaluation metrics for two different models on the same dataset and how using a time series model is not effective on static signs. This points us at a clear direction on how segregation of signs based on their movement can be useful when selecting the correct model for recognition. In our experiments, we were able to achieve an accuracy of approximately 85% for the train data, with the loss decreasing and leveling off after sixty to seventy iterations of training. This indicates that training for a sufficient number of iterations is crucial for achieving optimal results. We believe that our findings will contribute to further technological advancements in sign language recognition, ultimately improving communication and access for the deaf and hard-of-hearing community.

## 8. Future Scope

This research paper emphasizes the potential of technology-based solutions to improve communication and accessibility for the deaf and hard-of-hearing community in India, focusing on Indian Sign Language. However, there is still much work to be done to develop and deploy these technologies so that they can satisfy the needs of this community. Future research could examine the use of augmented reality (AR) and virtual reality (VR) technologies to enhance the acquisition and use of Indian Sign Language. AR and VR could provide a more immersive and interactive experience for users, allowing them to practice and improve their sign language skills in a more engaging and effective manner.

Future research could also focus on improving the accuracy and dependability of sign language recognition systems, particularly for Indian Sign Language, which has its own grammar and lexicon. This may necessitate the development of new machine learning algorithms tailored to Indian Sign Language, as well as the improvement of data capture and annotation procedures to ensure that systems are trained on a diverse and representative set of signs. In addition, the research could investigate the social and cultural factors that influence deaf and hard-of-hearing communities in India's adoption and utilization of sign language recognition systems. This could entail investigating the attitudes and perceptions of users toward these technologies, as well as the impact of societal and cultural barriers on their accessibility and effectiveness.

Overall, sustained research and development in the field of technology-based solutions for Indian Sign Language have the potential to significantly improve the way these people live life and accessibility of India's deaf population.

## 9. Conclusion

In conclusion, our research paper illuminates the ignored aspect of the DHH community in terms of technological advances. While technology has played a significant role in facilitating communication, education, and daily life for many individuals, the majority of existing technological solutions were not designed with the DHH community in mind. This lack of accessibility poses a substantial barrier to their complete integration into society.
It is our duty as a responsible society to ensure that everyone has equal access to technology. This requires developing technology that is tailored to the requirements of the DHH community, as opposed to focusing solely on solutions that make life easier for "normal" individuals.

Our research paper addresses this issue by focusing on the creation of technology that is tailored to this community's requirements. We have discussed the two categories of signs used in sign language - static and gesture. In order to circumvent these constraints, we employed LSTM, which has proven effective at encoding temporal dependencies in sequences, whereas CNN proved to be effective for static signs. Thus, we have contributed to the development of more precise and effective models for SLR.
Finally, we emphasize the significance of developing systems that facilitate two-way communication. Our research paper contains a section on text/speech-to-sign language conversion, which is a crucial stage in the development of systems that support both modes of communication. By doing so, we can guarantee that the DHH community will not be left behind in the digital era and will be able to fully participate in society.